\documentclass[10pt, conference]{IEEEtran}

\usepackage{graphicx} 
\usepackage{float}
\usepackage{cite}
\usepackage{amsmath,amssymb,amsfonts}
\usepackage{algorithmic}
\usepackage{textcomp}
\usepackage{pgfplots}
\usepackage{booktabs}
\usepackage{subcaption}
\usepackage{comment}
\pgfplotsset{compat=1.17}
\usepackage{array}
\usepackage{multicol}
\usepackage{titlesec}
\title{CogDrift-4D: Hierarchical Multimodal Modeling of Temporal Behavioral Decay for MCI Detection}
\author{\IEEEauthorblockN{ 
Sankalp Vaish$^1$,
Sumukh Veeramalla$^1$,
Rishank Singh$^1$,
Caroline Summerour$^2$,
John A.  Batsis$^2$,
Xiaohui Liang$^1$
}

		\IEEEauthorblockA{$^1$Department of Computer Science,
	University of Massachusetts Boston, MA, USA
    }
    \IEEEauthorblockA{$^2$School of Medicine,
	University of North Carolina at Chapel hill, NC, USA
		}
		Email: \{Sankalp.Vaish001, Sumukh.Veeramalla001, Rishank.Singh001, Xiaohui.Liang\}@umb.edu, \\Caroline\_Summerour@med.unc.edu, John.Batsis@unc.edu
}
\date{September 2025}

\begin{document}

\maketitle

\begin{center}
\footnotesize
\copyright~2026 IEEE. Personal use of this material
is permitted. Permission from IEEE must be obtained
for all other uses, in any current or future media,
including reprinting/republishing this material for
advertising or promotional purposes, creating new
collective works, for resale or redistribution to
servers or lists, or reuse of any copyrighted
component of this work in other works.

\vspace{0.5em}

\textit{Accepted for publication at the
2026 IEEE International Conference on E-health
Networking, Application \& Services (Healthcom).}
\end{center}

\begin{abstract}
Early detection of mild cognitive impairment remains a critical bottleneck in dementia care, demanding scalable tools capable of capturing subtle behavioral decay.
To capture the subtle temporal changes of cognitive decline, we introduce a behavioral dataset collected across seven unscripted multimodal interactions over an 18-month horizon. This dataset formalizes continuous, multi-session interactions into a dense 4D tensor ($\text{Batch} \times \text{Session} \times \text{Time} \times \text{Feature}$) that explicitly preserves concurrent local dynamics alongside long-term changes. To effectively process these complex behavioral patterns, we propose \textbf{CogDrift-4D}: a novel, hierarchical 4D Transformer framework.
To capture the complex trajectory of cognitive decline, CogDrift-4D implements a dual-level behavioral fatigue and drift modeling architecture. At the intra-session level, the model computes localized least-squares regression slopes to quantify the acute ``fading'' of frame-level facial action units and acoustic energy. At the inter-session level, a recursive cross-attention feedback encoder acts as a cumulative fatigue accumulator, mapping the patient's continuous behavioral drift across successive sessions. To mitigate environmental noise and track failures across sessions, a localized Quality Gate dynamically filters low-integrity features before final fusion, also an early fusion that integrates semantic insights with raw physical dynamics.
Evaluated on structured verbal and linguistic interactions, 
our architecture successfully detects MCI with a classification accuracy of $0.733 \pm 0.13$ and an F1-score of $0.716 \pm 0.14$, outperforming advanced baselines.
 The framework provides a mathematically rigorous, scalable, and robust foundation for home-based, real-time dementia screening.
\end{abstract}
\vspace{-1.5mm}
\section{Introduction}
Dementia is a major global public health challenge whose prevalence is rising with aging populations. Mild Cognitive Impairment (MCI) marks the transitional stage between normal age-related decline and early dementia, involving measurable deficits in memory, language, or executive function while daily independence is largely preserved.
Individuals diagnosed with MCI are at increased risk of developing Alzheimer's disease, with studies estimating that approximately 10\% to 20\% of adults aged 65 or older with MCI transition to dementia annually~\cite{alzheimers_mci}.
However, standard early screening tools such as the MoCA and MMSE depend on in-person clinical settings and trained administrators, limiting accessibility, longitudinal monitoring, and scalable early screening.

Motivated by the need for longitudinal monitoring approaches, we investigated the feasibility of tracking MCI progression by modeling long-term behavioral dynamics over an 18-month longitudinal horizon, in which participants completed standardized speech interaction tasks at consecutive 3-month intervals (7 sessions). 
Leveraging the continuous nature of human expression, this study captures physical facial dynamics, acoustic characteristics, and linguistic patterns. Fine-grained facial affect signals~\cite{Face1} and facial Action Unit (AU) activations~\cite{AU1} are extracted from video recordings to identify variations between cognitively healthy individuals and those with MCI. 
We track these behavioral and temporal signatures across successive sessions to evaluate longitudinal decline.



In recent years, several techniques~\cite{jmir2026, mdpi2026, Mbehrt, AMLA} have proven highly successful in extracting localized diagnostic markers from isolated, single-session recordings of spontaneous speech, facial micro-expressions, and linguistic features~\cite{pmc12631619}. However, most paradigms fail to capture long-term changes because they are restricted to static snapshots or flat sequence processing.
Frame-level facial landmarks, acoustic features~\cite{Speech1}, and text embeddings~\cite{Sent1, Sent2} exhibit substantial temporal variation, multi-scale dynamics, and complex cross-modal dependencies that conventional mean pooling or static classifiers cannot adequately model.
To address these data-specific limitations, our pipeline introduces a context-aware dual-stage fusion to resolve multi-resolution temporal dynamics, adaptive quality gating to dynamically suppress low-integrity or missing sessions, and localized least-squares regression to explicitly isolate vocal fatigue from true cognitive drift (gradual trajectory of decline in a patient's cognitive functions).
Our dual-stage fusion architecture first performs context-aware early fusion by integrating sentence embeddings with acoustic and expressive features within each session. Following session processing, a secondary inter-session fusion stage processes cross-modal representations over the 18 months (7 sessions) to identify cognitive decline.

The goal of this research is to lay the groundwork for an early-warning home-based, non-invasive cognitive screening ecosystem capable of operating robustly within unconstrained, everyday environments.
The system is highly adaptable, allowing for seamless integration into existing telehealth platforms, smart home virtual assistants, and mobile health applications. Unlike prior diagnostic methodologies that restrict their scope to isolated, single-session evaluations, our framework synthesizes longitudinal interactions, capable of tracking parallel cognitive and behavioral fluctuations over time.
This work makes the following key contributions:

\begin{itemize}


\item \textbf{Hierarchical 4D Transformer (CogDrift-4D):} We introduce a novel, end-to-end hierarchical architecture designed specifically to capture multi-scale temporal degradation in behavioral tracking data. The model operates on unified four-dimensional spatiotemporal tensors ($\text{Batch} \times \text{Session} \times \text{Time} \times \text{Feature}$) generated by a dedicated multi-modal alignment pipeline. 
The model leverages lower-level intra-session attention blocks to align asynchronous multi-modal features, coupled with an upper-level inter-session encoder to track macro-level longitudinal trajectory. CogDrift-4D
achieves a peak classification accuracy of 0.733 ( Outperforms advanced benchmarks by $\sim$8\%).


\item \textbf{A Longitudinal Conversational Dataset:} We introduce a novel, multi-session longitudinal behavioral dataset collected using Amazon Alexa. Moving beyond traditional clinical datasets that rely on rigid, highly structured laboratory tests, our tasks are explicitly designed for high ecological validity, capturing unscripted natural conversations alongside diverse cognitive behavioral tasks.

\item \textbf{Temporal analysis across seven sessions.}  
We examine temporal stability in emotional expressions and behavioral cues across repeated sessions. MCI participants exhibit greater acoustic variability, disrupted timing, reduced vocal stability, reduced emotional diversity, and persistent negative affect, while cognitively healthy participants show stable speech patterns, greater emotional 
variability, and more positive affect. 

\end{itemize}

\section{Related Work}


Conventional dementia screening depends on standardized clinical assessments administered by trained professionals in controlled settings, limiting repeated testing and large-scale deployment. This has increased interest in automated, contactless screening methods, with prior work largely focusing on speech-based biomarkers such as acoustic, prosodic, and linguistic features from conversational speech ~\cite{Speech1,Speech2,RI2}, as well as transcript-derived linguistic cues [~\cite{AudioTrans1}]. While these approaches demonstrate the value of speech for cognitive assessment, they often overlook nonverbal behavioral signals. Facial expressions and muscle activations provide complementary cognitive-health information, and prior studies suggest that facial cues can distinguish individuals with dementia~\cite{Face1}. 

Recent progress in affective computing and sentiment analysis shows that multimodal behavioral cues can improve machine learning-based interpretation of human states~\cite{Sent1,Sent2}. Although LLMs benefit from domain-specific pretraining, fine-tuning, and prompt engineering~\cite{LLMH1,LLMH3,LLMH2}, their computational cost and text-centered design make them less suitable for direct analysis of continuous audio and video streams. Neuroimaging methods such as MRI and PET have also been studied for dementia diagnosis ~\cite{brain1,brain2}, but their cost, limited accessibility, and clinical dependence restrict their use for home-based longitudinal monitoring. In contrast, everyday behavioral signals from speech, facial expressions, and language can be captured unobtrusively through consumer devices. Building on Cog-TiPRO~\cite{CogTiPRO} and prior work~\cite{RI3}, our study extends voice-assistant-based cognitive monitoring by incorporating a multimodal Transformer-based pipeline.
\section{Methodology}
\subsection{Preprocessing}


The dataset consisted of synchronized audio–video recordings of participant interactions with Alexa. All modalities were preprocessed to extract meaningful behavioral features as:

\textbf{Video Encoder:}
To address the presence of multiple individuals in each recording (participant, interviewer, Alexa device, questionnaire holder), we first applied MiVOLO ~\cite{MiVOLO} to every frame for face detection and age estimation. This allowed us to isolate the older adult participant across all sessions. Video models are trained using facial Action Units or emotion features. 
We employed ResEmoteNet ~\cite{ResEmoteNet}, to estimate frame-level emotional states (happy, sad, angry, surprise, neutral, disgust, fear). For each speaking segment, we aggregated these predictions into quantitative descriptors ~\cite{Face1}.
To characterize detailed facial behavior, we used OpenFace~\cite{OpenFace1} to extract Facial Action Units (AUs), which encode the activation of individual facial muscles (e.g., AU1: Inner Brow Raiser, AU12: Lip Corner Puller). As physiologically grounded descriptors of facial movement~\cite{AU1}, AUs can capture subtle expression changes associated with cognitive decline.

\textbf{Audio Encoder}
Using Parselmouth and Librosa, participant speech segments were first isolated and subsequently analyzed to extract both acoustic and prosodic characteristics~\cite{AudioAn1,AudioAn2}. Acoustic features included MFCC statistics, harmonics-to-noise ratio, etc., while prosodic and speech biomarkers comprised fundamental frequency (F0) measures, intensity contours, jitter, shimmer, speaking rate, pause count, and articulation rate. These features were then encoded into fixed-length embeddings for downstream multimodal modeling.

\textbf{Text Encoder:}
Transcripts generated using Rev.ai were encoded 
into embeddings using a Sentence Transformer model.


\begin{figure}[ht]
    \centering
    \includegraphics[width=1\linewidth]{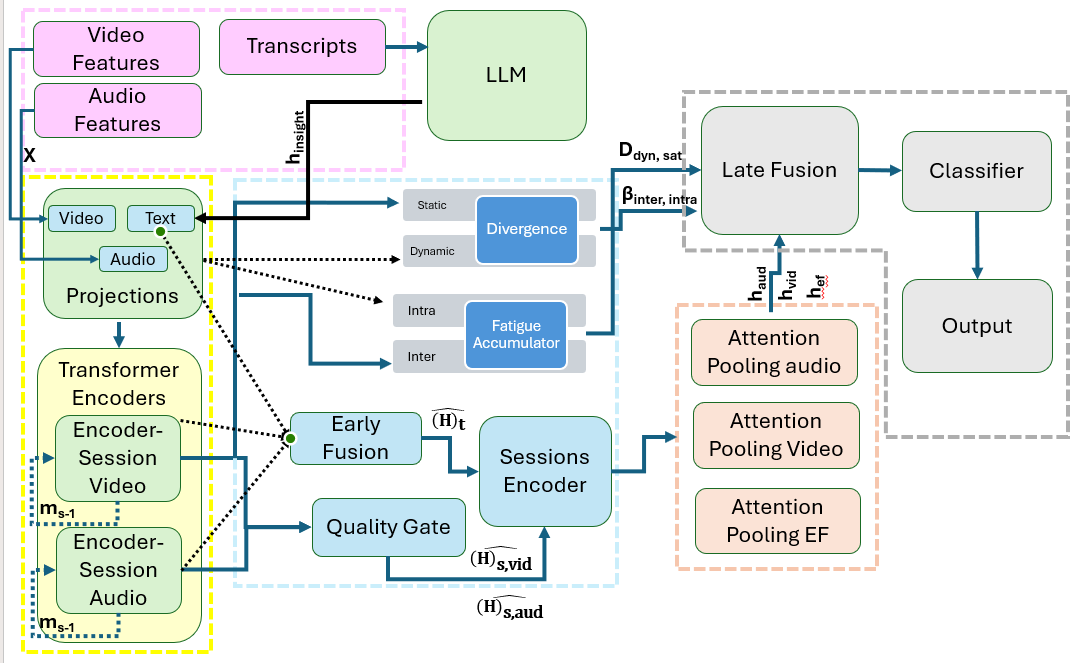}
    \caption{CogDrift-4D Framework}
    \label{fig: Model Pipeline}
\end{figure}

\subsection{\textbf{CogDrift-4D Framework}} 
The CogDrift-4D architecture is designed to capture the longitudinal decay of behavioral signals across multiple assessment tasks. The model processes data through a hierarchical pipeline of temporal encoding, divergence modeling, and multi-session fusion.
Figure \ref{fig: Model Pipeline} presents the proposed Transformer-based Multimodal architecture, using extracted features\cite{TS1, TS2, RI}.

\subsubsection{Intra-Session Temporal Encoding}
Each session is first processed at the frame level to capture fine-grained behavioral dynamics.

\textbf{Feedback-Driven Session Encoder}
To model the longitudinal progression across the $S$ assessment tasks, we employ a recursive feedback mechanism. Given the position-aware input tensor $\mathcal{X} \in \mathbb{R}^{B \times S \times T \times D}$, let $\mathbf{m}_{s-1} \in \mathbb{R}^D$ represent the memory state (summary) of the preceding session. The current session features $\mathbf{H}_s$ are updated via a cross-attention operation, where $\mathbf{H}_s$ acts as the query and $\mathbf{m}_{s-1}$ provides the contextual key and value:
\begin{equation}
\hat{\mathbf{H}}_s = \text{MultiHeadAttn}(Q=\mathbf{H}_s, K=\mathbf{m}_{s-1}, V=\mathbf{m}_{s-1})
\end{equation}
This architecture allows the model to utilize the cumulative "behavioral history" of the participant as a prior for interpreting the current task, effectively capturing inter-session fatigue and cognitive fluctuations. 
We apply a Transformer encoder where \textbf{Mask-aware attention} ensures the model attends only to valid frames $\mathcal{V}$ within each session:
\begin{equation}
\text{Attn}(Q, K, V, \mathbf{M}) = \text{softmax}\left( \frac{QK^\top}{\sqrt{d_k}} + \mathbf{M} \right)V
\end{equation}

where $\mathbf{M}$ is a binary mask preventing information leakage from padding. This process yields the modality-specific enriched representations $\hat{\mathbf{H}}_{s,\text{aud}}$ and $\hat{\mathbf{H}}_{s,\text{vid}}$.

\textbf{Dynamic Divergence:} To capture the ``rhythm'' of talking, we compute finite differences at multiple temporal offsets $\tau \in \{1,2,4\}$. For a feature sequence $\mathbf{X}$, the delta is defined as:
$
\Delta_\tau \mathbf{X}_t = \mathbf{X}_t - \mathbf{X}_{t-\tau}
$
The original signal is augmented with these multi-scale deltas. We define the Dynamic Divergence Score as the cosine similarity between the encoded temporal signatures of the visual and acoustic streams:

$
\mathcal{D}_{dyn} = \cos\left( \text{Enc}(\Delta \mathbf{X}_{vid}), \text{Enc}(\Delta \mathbf{X}_{aud}) \right)
$

\textbf{Static Divergence:} While dynamic divergence tracks frame-to-frame fluctuations, Static Divergence identifies persistent affective mismatches throughout the session. Using the attention-pooled session descriptors $\mathbf{h}_{vid}$ and $\mathbf{h}_{aud}$, the score is computed as the complement of their similarity:

$
\mathcal{D}_{stat} = 1 - \cos\left( \mathbf{h}_{vid}, \mathbf{h}_{aud} \right)
$

Dynamic (Rhythm): The patient’s face and voice are "out of sync" in time (e.g., the smile comes 1 second after the laugh).
Static (State): The patient’s face and voice are "out of sync" in emotion (e.g., the face looks sad, but the voice sounds angry).

\subsubsection{Neuro-Symbolic Insights}
By utilizing a sentence transformer model to directly process raw transcripts, we map textual inputs into dense, context-aware semantic embeddings. These embeddings serve as high-level linguistic anchors within our early-fusion layer, allowing the network to jointly correlate subtle shifts in word choice and conversational structure with concurrent acoustic and expressive features.
$
\mathbf{h}_{insight} = \text{SentenceTansformer}(\text{Transcript})
$

\subsubsection{Context-Aware Early Fusion and Quality Gating}
To bridge the gap between cognitive intent and multi-modal behaviors, we implement a context-aware early fusion layer. 
We enrich the frame-level physical representations by projecting the concatenated multimodal features into a shared latent space $\mathbf{H}_{t} \in \{V_t, A_t\}$ and $H_{in} \in \mathrm{Proj}(h_{\text{insight}})$:
\begin{equation}
\tilde{\mathbf{H}}_t = \text{ReLU}\left( W_f [ \mathbf{H}_{t} ; \mathbf{H}_{in} ] + b_f \right)
\end{equation}

This joint modeling allows the architecture to learn cross-modal temporal dependencies, where the linguistic context acts as a semantic prior for interpreting facial and acoustic dynamics. Concurrently, to safeguard against noisy or missing streams, we introduce a session-level Quality Gate mechanism $\mathbf{G}_s$ that dynamically assesses the signal reliability of each input channel:
$\mathbf{G}_s = \sigma(W_g \mathbf{h}_s + b_g)$
where $\mathbf{h}_s$ represents the aggregate session-level representation, and $\mathbf{G}_s = [g_{\text{vid}}, g_{\text{aud}}, g_{\text{fused}}] \in \mathbb{R}^3$ denotes the channel-specific gating coefficients bounded between 0 and 1 via the sigmoid ($\sigma$) activation. These coefficients are then applied element-wise to scale the respective representations:

\begin{align}
\hat{\mathbf{H}}_{s,\text{vid}} &= g_{\text{vid}} \cdot \mathbf{H}_{s,\text{vid}}, \\
\hat{\mathbf{H}}_{s,\text{aud}} &= g_{\text{aud}} \cdot \mathbf{H}_{s,\text{aud}}, \\
\hat{\mathbf{H}}_t &= g_{\text{fused}} \cdot \tilde{\mathbf{H}}_t.
\end{align}

This scalar acts as a soft filter; sessions identified as "noisy" or "uninformative" are scaled toward zero, ensuring that the model's diagnostic decision is grounded primarily in high-fidelity behavioral data.
\subsubsection{Behavioral Drift \& Fatigue Modeling}
A primary innovation of CogDrift-4D is the quantification of longitudinal performance decay. We model fatigue as a multi-level hierarchical process to capture both acute, task-specific decay and cumulative cognitive exhaustion.

\textbf{Level 1: Intra-Session Decay.} For each session $s$, we quantify the "fading" of behavioral intensity by calculating the least-squares regression slope $\beta$ of the feature magnitude. Let $y_t = \|\mathbf{h}_t\|_2$ denote the $L_2$ norm of the latent features at frame $t$. The slope is computed against the normalized time indices $\tau \in [0, 1]$ as follows:
\begin{equation}
\beta = \frac{\sum_{t=1}^{T} (\tau_t - \bar{\tau})(y_t - \bar{y})}{\sum_{t=1}^{T} (\tau_t - \bar{\tau})^2}
\end{equation}
This scalar serves as a proxy for acute exhaustion, where a negative $\beta$ indicates a significant decrease in expressive or acoustic energy during a single assessment task.

\textbf{Level 2: Inter-Session Accumulation.} To track global clinical progression across the interview, we aggregate intra-session descriptors via a recursive feedback loop. Let $\mathbf{A}_s$ be the accumulated fatigue state at session $s$, defined as:
$
\mathbf{A}_s = \text{GRU}(\mathbf{h}_s, \mathbf{A}_{s-1})
$
where the hidden state $\mathbf{A}$ serves as a \textit{Cumulative Fatigue Accumulator}. By comparing states $\mathbf{A}_1$ through $\mathbf{A}_S$, the model quantifies the inter-session "behavioral drift," identifying patients whose performance degrades holistically over the duration of the 4D assessment.

\subsubsection{Hierarchical Global Attention and Pooling}
Following temporal encoding, the modality-specific sequences $\hat{\mathbf{H}}_{s,\text{vid}}$, $\hat{\mathbf{H}}_{s,\text{aud}}$, and the context-aware temporal stream $\hat{\mathbf{H}}_t$ are processed by a \textbf{Global Encoder}. This module captures high-level dependencies across modalities before aggregating the data into session-level descriptors. To transition from 4D frame-level tensors to a unified representation, we employ a multi-branch attention pooling mechanism.

For each stream $k \in \{\text{vid, aud, temp}\}$, we compute scalar importance scores $\alpha_t^k$ via a learned linear projection $\mathbf{w}_a$. The normalized weights $\tilde{\alpha}_t^k$ are obtained via a softmax operation over the set of valid (non-padded) frames $\mathcal{V}$:
\begin{equation}
\alpha_t^k = \mathbf{w}_a^\top \hat{\mathbf{H}}_{k, t}, \quad \tilde{\alpha}_t^k = \frac{\exp(\alpha_t^k)}{\sum_{i \in \mathcal{V}} \exp(\alpha_i^k)}
\end{equation}

The final session-level embeddings, $\mathbf{h}_{\text{vid}}$, $\mathbf{h}_{\text{aud}}$, and $\mathbf{h}_{\text{ef}}$, are computed as the attention-weighted sum of their respective frame-level representations:
$
\mathbf{h}_k = \sum_{t \in \mathcal{V}} \tilde{\alpha}_t^k \hat{\mathbf{H}}_{k, t}
$


\subsubsection{ Late Fusion}
The final patient descriptor $\mathbf{h}_{final}$ is constructed via late fusion, integrating all specialized heads:
$
\mathbf{h}_{final} = [ \mathbf{h}_{vid}; \mathbf{h}_{aud}; \mathbf{h}_{ef}; \mathcal{D}_{dyn, stat}; \beta_{intra, inter} ] \odot \mathbf{G}_s
$
The final output $\hat{y}$ is obtained via a linear classifier:
$
\hat{y} = \sigma(W_{out} \mathbf{h}_{final} + b)
$

\textbf{Network Depth and Regularization:}
The SessionTransformer employs a hidden dimension of $d=128$ and consists of $L=8$ Transformer encoder layers with $H=8$ attention heads per layer. 
Dropout with rate $p=0.3$ is applied with the AdamW optimizer with a learning rate of $3\times10^{-5}$, weight decay of $1\times10^{-4}$ and cross-entropy loss for 50 epochs.

\section{Dataset}

The dataset was collected from 35 older adult participants through a series of structured interactions with Amazon Alexa. Each experimental session was divided into three distinct tasks, designed to elicit different forms of verbal and cognitive engagement.
These included question responses, participant-generated queries, and system-prompted answering scenarios, collectively capturing variations in language production, comprehension, and recall. To capture longitudinal variations, the entire experiment was repeated seven times with each participant at different time intervals. All sessions were recorded with synchronized audio and video, and transcripts were generated for subsequent analysis. 
In total, this resulted in (35 $*$ 3 = 105) data points, each sample is structured as a 4-D tensor spanning up to 7 sessions containing multiple intra-session time frames.
Data completeness was preserved, with the exception of 2 participants who completed only 6 sessions.

\begin{figure}[ht]
    \centering
    \includegraphics[width=1\linewidth]{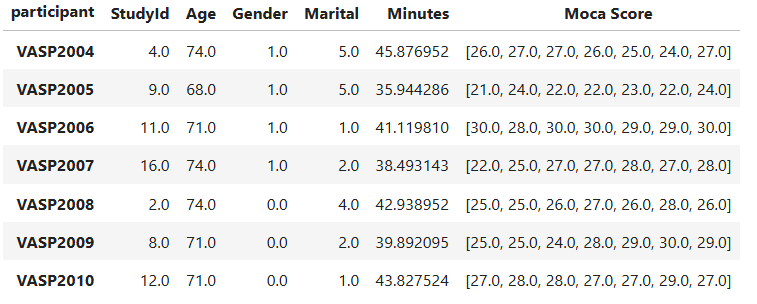}
    \caption{Data}
    \label{fig: Data}
\end{figure}

Figure \ref{fig: Data}  provides an overview of the dataset structure, illustrating the duration and composition of each experimental session. Each session lasted approximately 40–75 minutes and included all three activities. Participants with MoCA scores between 18 and 25 were categorized as having Mild Cognitive Impairment (MCI), while those scoring 26 or above were considered cognitively healthy. 
The figure lists all participant IDs and their corresponding session numbers. No personal information is included, ensuring strict privacy compliance.


\section{Results}

\subsection{Session Analysis}

\begin{figure}[ht]
    \centering
    \includegraphics[width=1\linewidth]{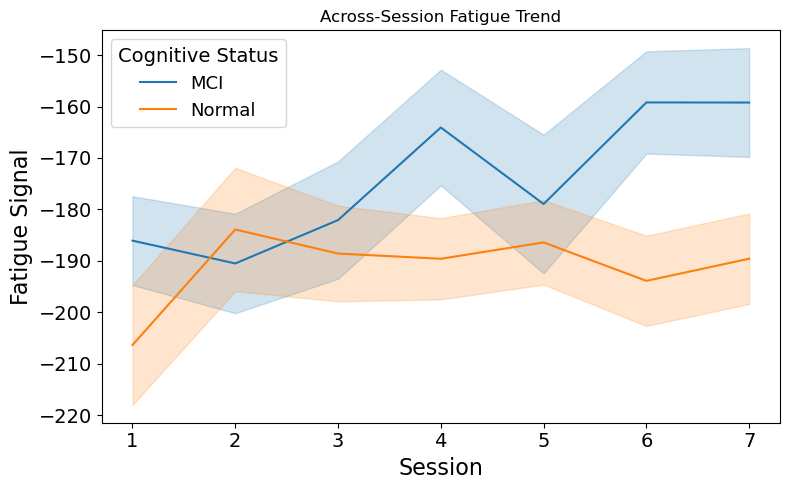}
    \caption{Fatigue across sessions}
    \label{fig: Fatigue across sessions}
\end{figure}

Figure~\ref{fig: Fatigue across sessions} shows the temporal fatigue variation across seven recorded sessions for a participant diagnosed with MCI. To quantify user exhaustion, we derive an acoustic fatigue metric by capturing the relationship between vocal energy and temporal processing delays. 
The MCI (blue) dominates across sessions, reflecting a consistently higher fatigue, while the healthy (orange) have lower fatigue.


\begin{figure}[ht]
    \centering
    \includegraphics[width=1\linewidth]{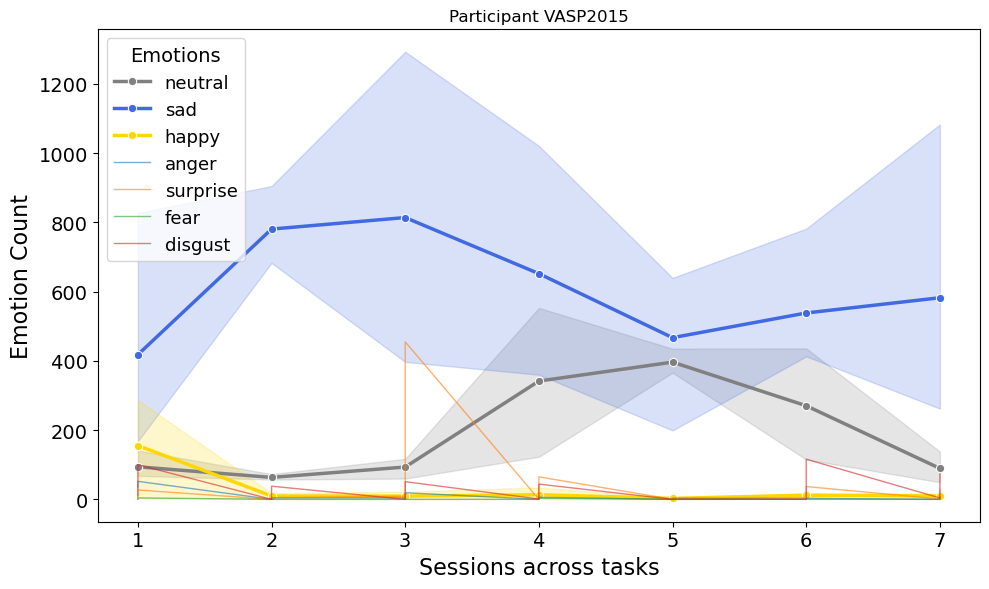}
    \caption{Emotions across sessions of a person with MCI}
    \label{fig: Emotions across sessions of a person with MCI}
\end{figure}


Figure~\ref{fig: Emotions across sessions of a person with MCI} shows the temporal variation of emotions across seven recorded sessions for a participant diagnosed with MCI. The sad emotion (blue) dominates across sessions, reflecting a consistently negative affective state, while the happy emotion (orange) appears infrequently, indicating reduced emotional variability and positivity.

\begin{figure}[ht]
    \centering
    \includegraphics[width=1\linewidth]{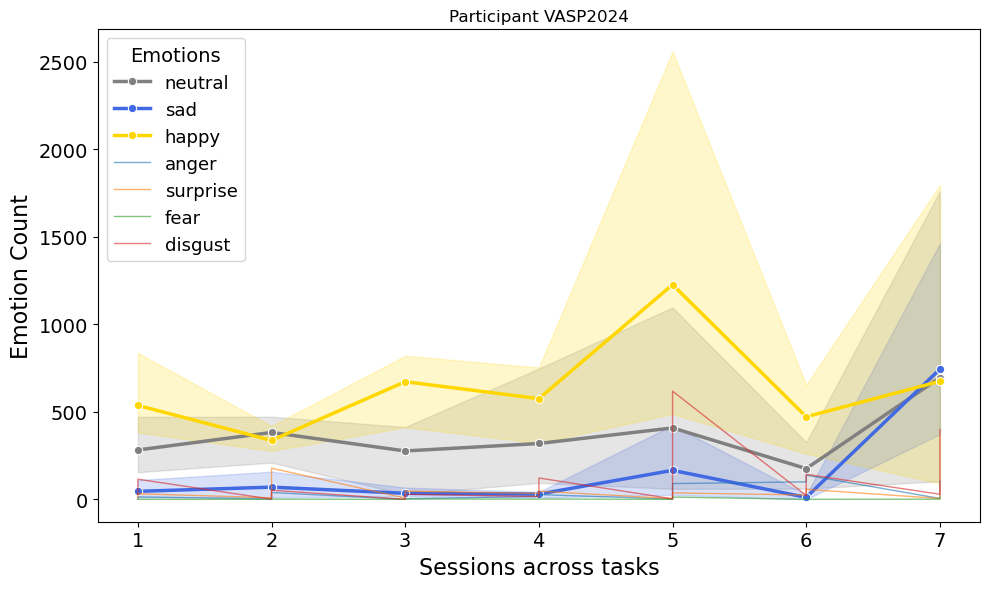}
    \caption{Emotions across sessions of a healthy person}
    \label{fig: Emotions across sessions of a normal person}
\end{figure}


Figure~\ref{fig: Emotions across sessions of a normal person} depicts emotion variation across seven sessions for a cognitively healthy participant. In contrast to the MCI case, happy (orange) and neutral (gray) emotions occur more frequently and consistently over time, while negative emotions, such as sadness, appear less often. This stability and predominance of positive affect suggest a potential link between emotional regulation and preserved cognitive functioning.

\subsection{Model Analysis}



Our CogDrift-4D framework was implemented in PyTorch and optimized on an NVIDIA RTX GPU.
We used binary classification, where $1$ denotes MCI and $0$ represents healthy controls, using the standard clinical threshold of MoCA $< 26$.
To establish a rigorous evaluation benchmark, three state-of-the-art clinical architectures were adapted to accept our uniform 4D behavioral tensors: \textbf{M-BEHRT}~\cite{Mbehrt}, which models consecutive tasks as sequential electronic health record visits; \textbf{AAMLA}~\cite{AMLA}, which utilizes a parameterized bilinear product for cross-modal affinity token alignment; and \textbf{PRA-PoE}~\cite{PRA-PoE}, which handles incomplete data streams using a closed-form product-of-experts rule.
To evaluate the specific impact of multi-scale temporal modeling, we implemented two internal structural baselines: \textbf{Baseline 1 (Static Late-Fusion MLP)} and \textbf{Baseline 2 (Isolated Multimodal Transformer)}. Baseline 1 compresses the entire 18-month timeline using mask-aware mean pooling to merge the longitudinal sessions into a single static array, fusing the modalities via concatenation before routing them through a multi-layer perceptron. Baseline 2 isolates each 3-month session, utilizing an intra-session transformer encoder to mix facial, acoustic, and linguistic tokens independently per task milestone. The resulting unaligned session representations are then globally averaged. 

\begin{table}[h]
\centering
\caption{Comparision study results (5-fold cross-validation, mean $\pm$ std).}
\label{tab:Comparision_study}
\begin{tabular}{lccc}
\hline
\textbf{Model Variant} & \textbf{Accuracy} & \textbf{F1-score} & \textbf{AUC} \\
\hline
Baseline 1             & 0.5429 $\pm$ 0.09 & 0.5429 $\pm$ 0.1 & 0.4990 $\pm$ 0.00 \\
Baseline 2        & 0.5524 $\pm$ 0.09 & 0.5413 $\pm$ 0.09 & 0.5528 $\pm$ 0.07 \\
AAMLA    & 0.5714 $\pm$ 0.07 & 0.5580 $\pm$ 0.06 & 0.5712 $\pm$ 0.04 \\
PRA-PoE          & 0.5810 $\pm$ 0.07 & 0.5665 $\pm$ 0.07 & 0.5588 $\pm$ 0.06 \\
M-BEHRT      & 0.6571 $\pm$ 0.13 & 0.6364 $\pm$ 0.17 & 0.6549$\pm$ 0.11 \\
CogDrift-4D            & \textbf{0.733 $\pm$ 0.13} &
                        \textbf{ 0.716 $\pm$ 0.15} &
                        \textbf{0.709 $\pm$ 0.14} \\
\hline
\end{tabular}
\end{table}
To prevent data leakage, we evaluate our proposed framework using a participant-grouped k-fold cross-validation, benchmarking performance across accuracy, F1-AUC, and ROC-AUC.
Table~\ref{tab:Comparision_study} presents the comparative evaluation using 5-fold cross-validation with 50 epochs of our proposed framework against the above architectures. 
In comparison, Baseline 1 and Baseline 2 yield near-chance accuracies of $0.5429 \pm 0.09$ and $0.5524 \pm 0.09$, respectively, confirming that flattening the 18-month longitudinal timeline or isolating single sessions obliterates the vital temporal indicators of cognitive decay. Furthermore, advanced public SOTA clinical benchmarks like AAMLA and PRA-PoE stall at ceiling accuracies of $0.5714 \pm 0.07$ and $0.5810 \pm 0.07$, demonstrating that optimization focused purely on single-session cross-modal alignment or probabilistic fusion cannot decipher long-term degradation. Crucially, even when measured against M-BEHRT—a powerful longitudinal sequence baseline that achieved a respectable accuracy of $0.6571 \pm 0.13$ and F1-score of $0.6549 \pm 0.11$, our framework yields a decisive $\sim$8\% improvement in overall accuracy. This proves that token-sequencing transformers are structurally ill-equipped to isolate the subtle cross-session behavioral drift and micro-fatigue slopes 
embedded within human expression and behavior, underscoring the necessity of CogDrift-4D's dual-stage design.

\begin{table}[h]
\centering
\caption{Ablation study results (10-fold cross-validation, mean $\pm$ std).}
\label{tab:ablation_study}
\begin{tabular}{lccc}
\hline
\textbf{Model Variant} & \textbf{Accuracy} & \textbf{F1-score} & \textbf{AUC} \\
\hline
Without Fatigue       & 0.714 $\pm$ 0.14 & 0.693 $\pm$ 0.16 & 0.687 $\pm$ 0.16 \\
Without Divergence    & 0.705 $\pm$ 0.14 & 0.689 $\pm$ 0.15 & 0.685 $\pm$ 0.14 \\
Without quality       & 0.722 $\pm$ 0.15 & 0.712 $\pm$ 0.15 & 0.703 $\pm$ 0.15 \\
Video + Audio + Text          & 0.705 $\pm$ 0.16 & 0.688 $\pm$ 0.18 & 0.681 $\pm$ 0.18 \\
Full Model            & \textbf{0.742 $\pm$ 0.17} &
                        \textbf{0.731 $\pm$ 0.18} &
                        \textbf{0.724 $\pm$ 0.18} \\
\hline
\end{tabular}
\end{table}


Table ~\ref{tab:ablation_study} evaluates the contribution of different components of the proposed multimodal framework using 10-fold cross-validation with 50 epochs. 
The full model achieved the best overall performance with an accuracy of 0.742, F1-score of 0.733, and AUC of 0.724, demonstrating the effectiveness of combining all proposed components.
Removing the fatigue-related features led to a reduction in performance across all metrics, indicating that fatigue cues provide important information for classification. Similarly, excluding the quality module slightly decreased performance, suggesting that quality-aware modeling contributes to more robust predictions. A significant performance drop was observed when the fatigue component was removed, with f1-score decreasing to 0.69 and AUC to 0.68, highlighting the importance of fatigue-based representations in capturing discriminative patterns. 



\section{Discussion}

The empirical findings of this study demonstrate that continuous, natural conversational behavior contains highly sensitive digital biomarkers capable of identifying early-stage cognitive impairment.
While our 18-month, 7-session longitudinal modeling framework successfully captured micro-fatigue slopes and macro-level behavioral drift, expanding the cohort to encompass a more prominent, multi-center participant demographic will be crucial to solidify the generalizability of these digital biomarkers against diverse geographic speech variations. 
We will also integrate lightweight LLMs to extract deep semantic and contextual insights directly from patient transcripts without compromising data privacy.
A promising future direction is to enable on-device deployment of the model, allowing privacy-preserving and low-latency cognitive assessment and supporting practical use in mHealth, telehealth, and smart-home monitoring applications.

\section{Conclusion}

This study introduces a fully non-invasive multimodal deep learning framework that integrates facial behavior dynamics, acoustic biomarkers, and linguistic features to distinguish cognitively healthy individuals from those experiencing MCI. By utilizing naturalistic interactions with Alexa, our proposed CogDrift-4D architecture successfully tracks fatigue, multi-scale temporal dynamics, and long-term longitudinal cognitive drift across an 18-month study.
CogDrift-4D achieves a peak classification accuracy of $0.733 \pm 0.13$ and an F1-score of $0.716 \pm 0.15$, yielding a performance margin ($\sim$8\% accuracy improvement) over advanced sequence benchmarks like M-BEHRT, and outperforming static late-fusion models by over 20\%. These results validate that early neurodegeneration manifests as a synchronized, multi-modal degradation across
expressive, acoustic, and linguistic channels that can be effectively captured via dual-stage hierarchical self-attention. Because the system operates entirely without specialized clinical instrumentation and requires only brief video recordings, it eliminates significant operational bottlenecks. Ultimately, this framework provides a highly scalable asset for seamless deployment within telehealth workflows, smart-home devices, and digital wellness platforms, enabling continuous, remote monitoring and low-cost early detection for cognitive decline.

\section{Acknowledgement}
This research was funded by the US National Institutes of Health National Institute on Aging R01AG067416.

\bibliographystyle{plain}

\bibliography{references}
\end{document}